\pdfoutput=1

\documentclass[11pt]{article}

\usepackage[]{acl}

\usepackage{times}
\usepackage{latexsym}

\usepackage[T1]{fontenc}

\usepackage[utf8]{inputenc}

\usepackage{microtype}

\usepackage{inconsolata}

\usepackage{graphicx}
\usepackage{amsfonts}
\usepackage{amssymb}
\usepackage{amsmath}
\usepackage{inconsolata}
\usepackage{caption}
\usepackage{longtable, array, booktabs}
\usepackage{authblk}

\title{Exploring Key Point Analysis with\\ Pairwise Generation and Graph Partitioning}

\author{\textbf{Xiao Li$^{\diamond}$, Yong Jiang$^{\dagger}$\textsuperscript{$\ast$}, Shen Huang$^{\dagger}$, Pengjun Xie$^{\dagger}$}\\
\textbf{Gong Cheng$^{\diamond}$\thanks{\hspace{1mm} Yong Jiang and Gong Cheng are the corresponding authors.}, Fei Huang$^{\dagger}$ } \\
 $^\diamond$State Key Laboratory for Novel Software Technology, Nanjing University \\
 $^\dagger$Institute for Intelligent Computing, Alibaba Group \\
  {\tt xiaoli.nju@smail.nju.edu.cn, gcheng@nju.edu.cn} \\
  {\tt \{yongjiang.jy,pangda,chengchen.xpj,f.huang\}@alibaba-inc.com} \\
}

\begin{document}
\maketitle
\begin{abstract}
Key Point Analysis (KPA), the summarization of multiple arguments into a concise collection of key points, continues to be a significant and unresolved issue within the field of argument mining.
\quad Existing models adapt a two-stage pipeline of clustering arguments or generating key points for argument clusters. This approach rely on semantic similarity instead of measuring the existence of shared key points among arguments.
Additionally, it only models the intra-cluster relationship among arguments, disregarding the inter-cluster relationship between arguments that do not share key points.
\quad To address these limitations, we propose a novel approach for KPA with pairwise generation and graph partitioning. 
Our objective is to train a generative model that can simultaneously provide a score indicating the presence of shared key point between a pair of arguments and generate the shared key point.
Subsequently, to map generated redundant key points to a concise set of key points, we proceed to construct an arguments graph by considering the arguments as vertices, the generated key points as edges, and the scores as edge weights. We then propose a graph partitioning algorithm to partition all arguments sharing the same key points to the same subgraph.  
Notably, our experimental findings demonstrate that our proposed model surpasses previous models when evaluated on both the ArgKP and QAM datasets.
\end{abstract}

\section{Introduction}
\begin{figure}[t]
    \centering
    \includegraphics[width=\linewidth]{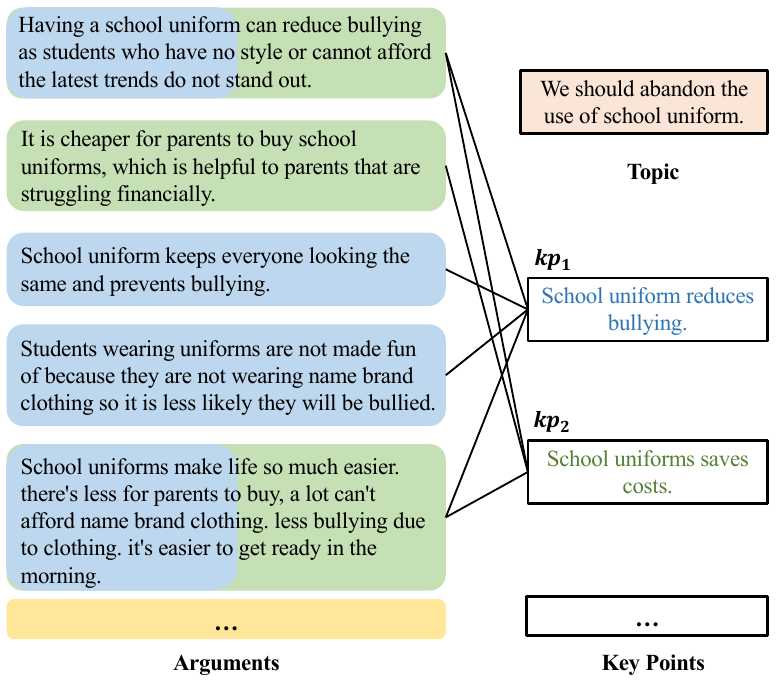}
    \caption{An example in the ArgKP dataset. Only a subset of arguments and key points are shown due to space limitation. Our goal is to generate key points~(right side) within a specific topic based on the provided arguments~(left side). A single argument may correspond to multiple key points.}
    \label{fig:sample}
\end{figure}

Analyzing a large number of arguments and making decisions is a long standing issue present in real life scenarios such as product review analysis~\cite{bar-haim-etal-2021-every,cattan-etal-2023-key}, opinion surveys~\cite{bar-haim-etal-2020-arguments}, etc. However, manually reading through a vast amount of arguments is not a feasible solution. In this context, key point analysis~(KPA, \citealp{bar-haim-etal-2020-arguments}) has been proposed. KPA involves distilling a multitude of arguments into a concise set of key points, followed by a quantitative evaluation, typically by calculating the frequency of each key point related to arguments. An example is shown in Figure~\ref{fig:sample}.

\paragraph{Existing Methods and Limitations.} The generation of key points from arguments represents the most pivotal and challenging aspect of the KPA task. Initially, a straightforward approach~\cite{cattan-etal-2023-key} involves selecting a few short arguments as key points and then determining whether other arguments match these key points. Subsequently, advanced model~\cite{li-etal-2023-hear} divides the KPA task into two stages. Firstly, a semantic feature based clustering model is utilized to cluster arguments into multiple clusters. Then, a generation model is employed to generate corresponding key points for arguments in each cluster by concatenating the arguments.
However, these existing methods possess certain limitations. 
\citealp{cattan-etal-2023-key} posits that the mere presence of appropriate key points within arguments is insufficient.
\citealp{li-etal-2023-hear}'s approach is based on clustering arguments using their semantic features and then positing that arguments within a cluster share the same key point. However, this approach does not fully align with the objectives of the KPA task. It is possible for arguments that are not semantically similar to still share a key point, as an argument can correspond to more than one key point. Moreover, \citealp{li-etal-2023-hear}'s approach only models the relationships between arguments that share a key point~(i.e., intra-cluster arguments) and neglects the relationships between arguments that do not share a key point~(i.e., inter-cluster arguments).

\paragraph{Our Approach.} 

In order to address the deficiencies mentioned above, we present an approach which trains a generative model to simultaneously utilize the relationships between intra-cluster and inter-cluster arguments. This model generates a key point for a pair of inter-cluster arguments and provided a score indicating whether there exists shared key points between the pair of arguments.
During the prediction phase, we utilize the score provided by the generative model, which indicates whether a pair of arguments is inter-cluster arguments, rather than the semantic similarity of the arguments, to partition the arguments into multiple clusters. This score is provided by a fine-tuned generative model, which is aligned with the data distribution within the domain, rather than employing a clustering model that is unrelated to the domain.
We then construct an argument graph and formulate the relationships among arguments as a graph-based problem. Specifically, we represent arguments as vertices in the graph, the shared key point between two arguments as edges, and the scores as edge weights, resulting in a weighted argument graph. We introduce an iterative graph partitioning algorithm, which generates several subgraphs, each representing a collection of arguments that share the same or similar key point. From these subgraphs, we select a representative key point for further analysis and utilization.

To summarize, our contributions include
\begin{itemize}
    \item proposing a pairwise generation approach that simultaneously utilizes the information between intra-cluster arguments and inter-cluster arguments and providing a score to indicate the presence of the shared key point between a pair of arguments, and 
    \item introducing a weighted graph partition algorithm designed specifically for the KPA task. Our algorithm evaluates the connections between arguments by the scores provided by the generative model, as semantic similarity may not fully align with the objectives of the KPA task.
\end{itemize}

\paragraph{Outline.} We elaborate our approach in Section~\ref{sec:approach}, present experiments in Section~\ref{sec:experiments}, discuss related work in Section~\ref{sec:related_work}, and conclude in Section~\ref{sec:conclusion}.

\paragraph{Code.} Our code and data are available on GitHub: \href{https://github.com/Alibaba-NLP/Key-Point-Analysis}{https://github.com/Alibaba-NLP/Key-Point-Analysis}.
\section{Approach}
\label{sec:approach}
On the KPA task, the input consists of a set of arguments $\mathbb{A}=\{(\texttt{arg}_1, \texttt{stance}_1), (\texttt{arg}_2, \texttt{stance}_2), \cdots\}$ under a given topic~$T$, where $\texttt{arg}_i$ is the argument text and $\texttt{stance}_i$ is the stance for $\texttt{arg}_i$. The stance~(``pro'' or ``con'') indicate the stance of argument towards the topic. The ideal outcome is a concise set of key points $\mathbb{K}=\{\texttt{kp}_1, \texttt{kp}_2, \cdots \}$, where each key point in $\mathbb{K}$ corresponds to one or more arguments in $\mathbb{A}$. That is, our objective is to learn a surjective function $f: \mathbb{A}, T\rightarrow \mathbb{K}$. Afterwards, the importance of each key point can be expressed by calculating the percentage of arguments corresponding to each key point.

\begin{figure}[t]
    \centering
    \includegraphics[width=0.9\linewidth]{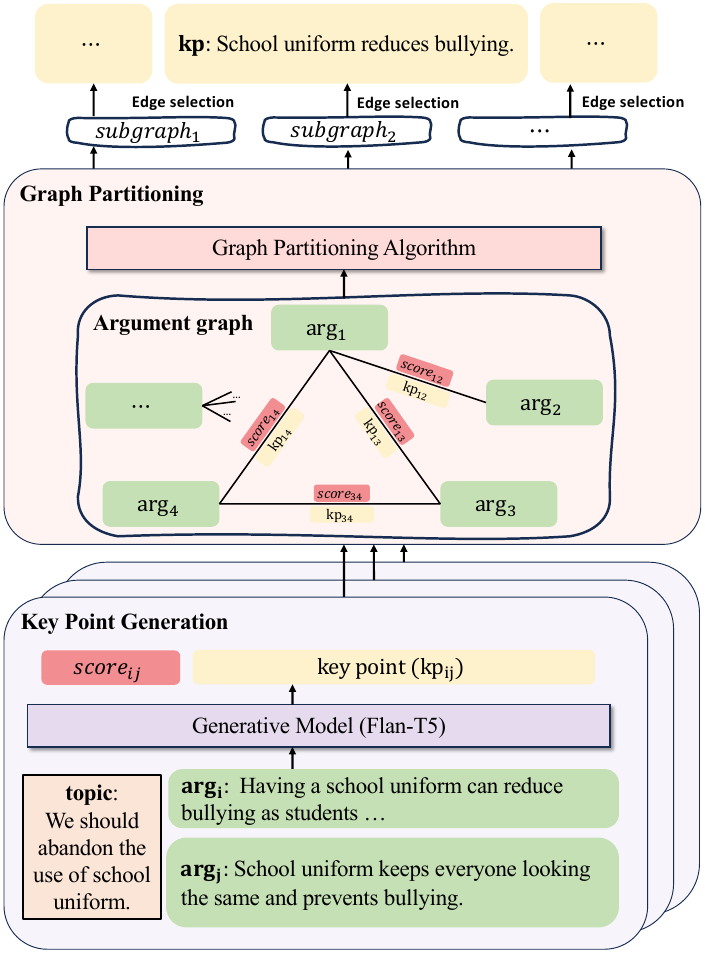}
    \caption{Overview of our approach. Each argument possesses a stance representing its polarity~(``pro'' or ``con''). For the sake of brevity, it has been omitted in the figure.}
    \label{fig:main}
\end{figure}

Figure~\ref{fig:main} provides an overview of our proposed method, which comprises two principal components. The initial component is dedicated to the generation of key points, as depicted at the bottom of Figure~\ref{fig:main}. During this phase, we process the given topic~$T$ and pairs of arguments~$\mathbb{A}\times\mathbb{A}$ to ascertain if each pair shares a key point and, if so, generate the corresponding key point. The subsequent component entails graph partitioning, illustrated at the top of Figure~\ref{fig:main}. In this stage, we construct a graph of arguments, key points, and their associated scores. We then apply our specially developed graph partitioning algorithm to divide the graph into multiple subgraphs, allowing for the possibility of the same argument being present in more than one subgraph, as a single argument may relate to multiple key points.
Finally, we select a representative key point from each subgraph to form the final concise set of key points $\mathbb{K}$. Notably, unlike previous pipeline approaches, we transmit the scores generated by the first stage to the second stage, avoiding issues such as feature propagation interruption and error accumulation that may arise in multi-stage pipelines.

\subsection{Key Point Generation}
\label{sec:kpg}
Inspired by contrastive learning~\cite{Khosla2020SupervisedCL,gao-etal-2021-simcse}, our approach takes a fine-grained perspective by pairing arguments and determining whether a pair of arguments shares the same key point.
We posit that it is equally important to model the relationships between inter-cluster arguments that possess different key points and intra-cluster arguments that share a key point.
The previous models either mapped each argument to a single key point~\cite{alshomary-etal-2021-key,kapadnis-etal-2021-team} or clustered arguments together and concatenated multiple arguments within a cluster to generate a key point~\cite{li-etal-2023-hear}.
\paragraph{Construction of Training Data.} We pack a pair of arguments into an input text
\begin{equation*}
    \texttt{input}_{ij}=\texttt{topic}\ |\ \texttt{stance}_i\ \texttt{arg}_i\ |\ \texttt{stance}_j\ \texttt{arg}_j,
\end{equation*}
and prefix the output text with a special word to indicate whether the input pair of arguments are intra-cluster arguments or inter-cluster arguments, which is 
\begin{equation*}
    \small
    \texttt{output}_{ij}=\begin{cases} 
             \text{Yes.}\ \{\texttt{kp}_{ij}\}, & \text{for intra-cluster arguments,} \\
             \text{No.},  & \text{for inter-cluster arguments.}
\end{cases}
\end{equation*}
We construct training data by pairing arguments, disregarding the order between a pair of arguments. That is, we consider $\texttt{input}_{ij}$ while discarding $\texttt{input}_{ji}$~(where $i<j$). We determine whether a pair of arguments are intra-cluster arguments or inter-cluster arguments by assessing if the corresponding key points of the arguments are identical. To avoid overfitting, we limit the occurrence of each argument as an intra-cluster argument in the training data to no more than five times.

\begin{table}[h]
\centering
\small
\resizebox{\linewidth}{!}{%
\begin{tabular}{|p{0.25\linewidth}p{0.7\linewidth}|}
\hline
\textbf{input A} & We should adopt atheism. | positive. if we adopt atheism then maybe people will start believing in the scientific community again. | positive. we should adopt atheism because science can explain how we got here without needing a god to explain it. \\
\hline
\textbf{output A} & Yes. Science can adequately explain the Universe \\
\hline
\hline
\textbf{input B} & We should ban the use of child actors. | positive. child actors do not get to live a proper young persons life and experience childhood play, education and social interaction, so should be banned | positive. the use of child actors should be banned as children are not capable of making important decisions and some do not have trustworthy parents to make rationale decisions. \\
\hline
\textbf{output B} & No. \\
\hline
\end{tabular}}
\caption{\label{table:training_examples}Training examples. The two arguments in example A share a key point, whereas the two arguments in example B do not.}
\end{table}

To facilitate a better understanding of the construction of the training data, we present two examples in Table~\ref{table:training_examples}.

\paragraph{Generation Model.} Subsequently, we feed $\texttt{input}_{ij}$ to the generative model~\footnote{In our paper, we exemplify our approach by employing a model based on the encoder-decoder architecture. However, our approach can also be applied to decoder-only models.},
\begin{equation*}
    \mathbf{h}_1, \mathbf{h}_2, \cdots, \mathbf{h}_n =\mathtt{Decoder}(\mathtt{Encoder}(\texttt{input}_{ij})),
\end{equation*}
where $\mathbf{h}_k$ denotes the representation obtained from the $\mathtt{Decoder}$ during the $k$-th decoding step and $n$ is the decoder output length. During the decoding phase, a linear transformation $\mathtt{LM\_Head}$ is employed to map the representations outputted by the $\mathtt{Decoder}$ to the probability distribution for the tokens in the vocabulary at step $k$, denoted as
\begin{equation*}
    S_k=\mathtt{LM\_Head}(\mathbf{h}_k).
\end{equation*}
The $k$-th token is decoded by
\begin{equation*}
    \texttt{token}_k=\mathtt{vocab}[\mathtt{argmax}(S_k)],
\end{equation*}
where $\mathtt{vocab}$ is the vocabulary.

We extract scores for token ``Yes'' and ``No'' from $S_1$~(i.e., $S_1[\text{Yes}]$ and $S_1[\text{No}]$, respesctively) and obtain the score indicating whether $\texttt{arg}_i$ and $\texttt{arg}_j$ sharing the same key point by
\begin{equation}
    score_{ij}=\frac{\mathtt{exp}(S_1[\text{Yes}])}{\mathtt{exp}(S_1[\text{Yes}])+\mathtt{exp}(S_1[\text{No}])}.
\label{eq:sharing_score}
\end{equation}
The loss during the training process is
\begin{equation*}
    \mathcal{L}=-\frac{1}{n}\sum_{1\le k\le n}\log\frac{\mathtt{exp}(S_k[\texttt{output}_{ij}[k]])}{\sum_{ t\in\texttt{vocab}}\mathtt{exp}(S_k[t])},
\end{equation*}
where $\texttt{output}_{ij}[k]$ is the $k$-th token in $\texttt{output}_{ij}$.

By employing this approach, we have integrated the binary classification task of determining the presence of shared key points and the task of generating key points into a single generative model, thereby mitigating potential information loss that may arise from coupling two different models.

\subsection{Graph Partitioning}
\label{sec:sgp}
After the generation of key points, it is necessary to aggregate all intra-cluster arguments and determine the shared key point. 
We transform this issue into a graph partitioning problem, such that each partition yields a subgraph representing a cluster of arguments that share key points.
Previous approaches~\cite{li-etal-2023-hear} relied on semantic similarity-based clustering methods. However, it is important to recognize that two arguments sharing the same key point may not always exhibit high semantic similarity, since an argument can encompass multiple key points. 
An alternative direct approach involves clustering the key points directly. However, the semantic gap between the generative models and the clustering models may introduce performance drop in this approach. We have mitigated this issue by transferring $score_{ij}$ between the two stages.
Specifically, in Equation~\ref{eq:sharing_score}, we calculate the score $score_{ij}$ to quantify the presence of a shared key point between argument $\mathtt{kp}_i$ and argument $\mathtt{kp}_j$. The score $score_{ij}$ is then used to guide the partitioning process.

\paragraph{Argument Graph Construction.}
We employ arguments~$\mathbb{A}$ as vertices and each edge possesses two attributes, an edge weight
\begin{equation*}
    score\in\mathbb{S}=\{score_{i,j}\lvert 1\leq i,j\leq \lvert \mathbb{A}\rvert, i\neq j \},
\end{equation*}
which indicates the probability of the presence of a shared key point, and the generated shared key point
\begin{equation*}
    \texttt{kp}\in\hat{\mathbb{K}}_{\text{Gen}}=\{\texttt{kp}_{ij}\lvert 1\leq i,j\leq \lvert \mathbb{A}\rvert, i<j \}.
\end{equation*}
We have removed the edges for which the model outputs ``No'', corresponding to the edges between argument vertices that do not share a key point.

In doing so, we construct a weighted argument graph
\begin{equation*}
    G=\langle \mathbb{A}, (\mathbb{S}, \hat{\mathbb{K}}_{\text{Gen}})\rangle.
\end{equation*}

\paragraph{Partitioning by Local Search.} We employ local search algorithm to partition argument graph. We use the BAAI-bge-large~\cite{Xiao2023CPackPR} model to obtain representations of arguments, and then apply k-means clustering~\cite{Lloyd1982LeastSQ} to cluster the arguments, which serves as the initial partitioning for our graph. We then obtain the initial subgraphs $\mathbb{G}_{\text{subgraph}}=\{g_1,g_2,\cdots,g_s\}$, where $s$ is a hyperparameter. Subsequently, we iteratively move vertices across the current subgraphs by optimizing a cost function, until no improvement of cost is possible or the limit of iteration steps denoted by $l$ is reached.
We finally obtain subgraphs $\mathbb{G}'_{\text{subgraph}}$ and form the concise set of key points $\mathbb{K}_{\text{Gen}}$ from each subgraph by choosing the key points with the highest edge weights.

\paragraph{Cost Function.} We define the weight $\mathtt{wt}(g)$ of a subgraph $g$ as the average weight of all edges in $g$. For a randomly selected subgraph and within it, a randomly selected vertex $\texttt{arg}$, we calculate the change in the graph weight resulting from the movement of vertex $\texttt{arg}$ from the subgraph $g_{\text{out}}$ to another subgraph $g$ by our cost function
\begin{equation*}
    \begin{aligned}
        \mathtt{cost}&(g, g_{\text{out}}, \texttt{arg})=\\
        &\mathtt{wt}(g_{\text{out}}\setminus \{\texttt{arg}\})-\mathtt{wt}(g_{\text{out}})+\\
        &\mathtt{wt}(g\cup\{\texttt{arg}\})-\mathtt{wt}(g).
    \end{aligned}
\end{equation*}
We then search for a target subgraph $g_{\text{in}}$ by
\begin{equation*}
    \begin{aligned}
        g_{\text{in}}=\mathop{\mathtt{arg\ max}}_{g\in \mathbb{G}_{\text{subgraph}}\setminus \{g_{\text{out}}\}} \mathtt{cost}(g, g_{\text{out}}, \texttt{arg}),
    \end{aligned}
\end{equation*}
which means that moving the selected vertex $\texttt{arg}$ from $g_{\text{out}}$ to $g_{\text{in}}$ would result in the maximum change in graph weight and the selected vertex will only be moved when the weight change is positive.

\paragraph{Soft Partition.} Furthermore, since an argument may encompass multiple key points, it is permissible to allow an argument to belong to multiple subgraphs simultaneously, which we refer to as soft partition. We set a threshold $h$, such that if the removal of the selected argument from its original subgraph would cause the weight of the subgraph to decrease beyond the threshold $h$, we will retain the argument within the original subgraph. This means that the selected argument can contribute a weight gain greater than $h$ in the original subgraph.

It is worth noting that we have not imposed any restrictions on the connectivity of subgraphs in our algorithm. This is because, given a specified number of subgraphs, we cannot guarantee that all the resulting subgraphs will be connected~(as the number of connected components in $G$ may be less than the number of subgraphs). However, it is important to note that our optimization objective~(i.e., graph weight) tends to favor connected subgraphs or subgraphs with higher connectivity.

\section{Experiments}
\label{sec:experiments}

\begin{table*}[t]
    \centering
    \resizebox{\textwidth}{!}{%
    \begin{tabular}{@{\extracolsep{5pt}}lrrrrrrrrrr@{}}
        \specialrule{1pt}{2pt}{2pt}
        & \multicolumn{5}{c}{ArgKP} & \multicolumn{5}{c}{QAM} \\
        \cline{2-6} \cline{7-11} 
        \textbf{Model} & \multicolumn{1}{c}{$\mathtt{Rouge}$-1} & \multicolumn{1}{c}{$\mathtt{Rouge}$-2} & \multicolumn{1}{c}{\textbf{sP}} & \multicolumn{1}{c}{\textbf{sR}} & \multicolumn{1}{c}{\textbf{sF1}} & \multicolumn{1}{c}{$\mathtt{Rouge}$-1} & \multicolumn{1}{c}{$\mathtt{Rouge}$-2} & \multicolumn{1}{c}{\textbf{sP}} & \multicolumn{1}{c}{\textbf{sR}} & \multicolumn{1}{c}{\textbf{sF1}} \\
        \specialrule{1pt}{2pt}{2pt}
        \multicolumn{11}{l}{\textbf{Previous SOTA Approaches}} \\
        GBS & 19.60 & 3.40 & 53.00 & 52.00 & 52.00 & - & - & - & - & - \\
        Enigma$_\text{PEGASUS}$ & 20.00 & 4.80 & 58.00 & 57.00 & 57.00 & \textbf{32.61} & 7.17 & 46.96 & \underline{44.43} & 45.51 \\
        SKPM$_\text{Flan-T5-base}$ & 30.30 & 8.90 & 59.00 & 58.00 & 59.00 & - & - & - & - & - \\
        SKPM$_\text{Flan-T5-large}$ & 31.40 & 9.10 & 57.00 & \textbf{62.00} & \underline{60.00} & - & - & - & - & - \\
        \specialrule{1pt}{2pt}{2pt}
        \multicolumn{11}{l}{\textbf{Ours}} \\
        w/ Flan-T5-base & \underline{37.38} & \underline{11.39} & \underline{61.29} & 58.08 & 59.59 & 30.31 & \underline{7.44} & \underline{48.62} & 43.81 & \underline{45.96} \\
        w/ Flan-T5-large & \textbf{50.91} & \textbf{21.23} & \textbf{62.57} & \underline{60.26} & \textbf{61.34} & \underline{30.44} & \textbf{8.02} & \textbf{49.38} & \textbf{45.27} & \textbf{47.11} \\
        \specialrule{1pt}{2pt}{2pt}
    \end{tabular}
    }
    \caption{Comparison with previous state-of-the-art approaches. The best results are in bold and the second best results are underlined. The dash signifies that we did not obtain the results due to issues such as code availability. The results of previous state-of-the-art approaches on ArgKP are taken from~\citealp{li-etal-2023-hear}. }
    \label{table:comparison_with_baselines}
\end{table*}

\subsection{Datasets}
We conducted experiments on ArgKP~\cite{friedman-etal-2021-overview} and QAM~\cite{guo-etal-2023-aqe}. ArgKP is a KPA dataset annotated by crowd sourcing, with training/validation/test sets consisting of 24/4/3 topics, 5583/932/724 arguments, and corresponding 207/36/33 key points. Each argument and key point possess a stance attribute, either ``pro''~(i.e., positive) or ``con''~(i.e., negative), indicating whether they support the current topic.
The QAM dataset is originally designed for extracting quadruples $\langle$key point, stance, argument, evidence type$\rangle$ from documents with a given topic. QAM is annotated through crowdsourcing, and we have transformed the QAM dataset into the format required for the KPA task. The training/validation/test sets consists 96/52/53 topics, 6635/808/948 arguments, and corresponding 2366/314/337 key points.

\subsection{Evaluation Metrics}
We employed $\mathtt{Rouge}$~\cite{lin-2004-rouge} and soft-Precision/Recall/F1 as metrics.
Following~\citealp{li-etal-2023-hear}, we obtained $\mathtt{Rouge}$ scores by comparing the concatenation of generated key points and the concatenation of reference key points. 
We reported the scores of $\mathtt{Rouge}$-1 and $\mathtt{Rouge}$-2. 
For soft-Precision/Recall/F1~(denoted by sP, sR and sF1, respectively), we employed $\mathtt{BLEURT}$~\cite{sellam2020bleurt} to measure the similarity between individual generated key point and the reference key point.
The sP, sR and sF1 to compare the similarity between the sets of key points were given by
\begin{equation*}
    \small
    \begin{aligned}
        \textbf{sP}&=\frac{1}{\lvert\mathbb{K}_{\text{Gen}} \rvert}\mathop{\sum}_{\texttt{kp}_\text{Gen}\in\mathbb{K}_{\text{Gen}}}\max_{\texttt{kp}\in\mathbb{K}}\mathtt{BLEURT}(\texttt{kp},\texttt{kp}_\text{Gen}), \\
        \textbf{sR}&=\frac{1}{\lvert\mathbb{K} \rvert}\sum_{\texttt{kp}\in\mathbb{K}}\max_{\texttt{kp}_\text{Gen}\in\mathbb{K}_\text{Gen}}\mathtt{BLEURT}(\texttt{kp},\texttt{kp}_\text{Gen})\ \text{and} \\
        \textbf{sF1}&=\frac{2\times \textbf{sP}\times \textbf{sR}}{\textbf{sP}+\textbf{sR}}.
    \end{aligned}
\end{equation*}

\subsection{Implement Details}
We conducted experiments on NVIDIA A100~(80GB) GPUs. Our implementations were based on PyTorch 2.0.1 and HuggingFace Transformers 4.30.2. We set $\textit{learning rate}=1e-5$ selected from $\{1e-5,3e-5,6e-5\}$, $\textit{batch size}=64$ selected from $\{32,64,128,256\}$, $\textit{random seed}=42$, $\textit{maximum sequence length} = 512$, $\textit{epoch}=5$. We set $h=0.008$ for ArgKP, $h=0.006$ for QAM and $l=200$ for ArgKP, $l=80$ for QAM. In order to ensure a fair comparison with previous work~\cite{li-etal-2023-hear}, we set the number of selected key points, which is the number of subgraphs $s$, to be equal to the number of reference key point set.

\subsection{Previous State-of-the-Art Approaches}
We compared with three state-of-the-art models: graph based summarization~(GBS, \citealp{alshomary-etal-2021-key}), Enigma~\cite{kapadnis-etal-2021-team} and supervised key point modelling~(SKPM, \citealp{li-etal-2023-hear}). The GBS model was a representative of the extractive key point generation models. It employed arguments to establish a graph, where each argument vertex was examined to determine their suitability as a key point. The PageRank~\cite{Page1999ThePC} algorithm was then utilized to choose the key points. Enigma generated a key point for each argument based on PEGASUS~\cite{10.5555/3524938.3525989} and selected the highest-scoring key point using the precision of $\mathtt{Rouge}$-1 compared with reference key point set. 
SKPM was a two-stage model that first clusterd arguments and then generated key points for arguments within each cluster.

\subsection{Comparison with Previous State-of-the-Art Approaches}
In comparison with the existing methods as shown in Table~\ref{table:comparison_with_baselines}, our approach demonstrated advantages over the previous SOTA approaches. 
Specifically, on the ArgKP dataset, we observed improvements of 19.51--31.31 points in $\mathtt{Rouge}$-1 and 1.34--9.34 points in sF1. On the QAM dataset, our approach yielded an improvement of 1.60 points in sF1. The results demonstrated the effectiveness of our method, which were analyzed in detail in the subsequent sections.

\subsection{Effectiveness of Key Point Generation}
We conducted several experiments to evaluate the effectiveness of our generative model in identifying and generating the shared key point between a pair of arguments. These experiments were structured around two research sub-questions. \textbf{SQ1}~(Shared Key Point Detection) focused on the ability of our model to determine the presence of a shared key point, which was essentially a binary classification task. \textbf{SQ2}~(Key Point Generation) assessed the model's capability to generate high-quality key points that effectively covered the reference key point set. 
In tackling these research sub-questions, we prioritized the recall metric because our subsequent step involves selecting a concise subset from an extensive set of key points, as explained in Section~\ref{sec:effectiveness_of_graph_partitioning}).

\subsubsection{SQ1: Shared Key Point Detection}
\label{sec:shared_key_point_detection}
We applied a representative model RoBERTa-large~\cite{Liu2019RoBERTaAR} to classify whether two arguments shared a key point and compared its performance with that of our generative model. In our generative model, if a key point was generated for a pair of arguments, the output was classified as positive~(i.e., the output began with ``Yes''). Conversely, if no key point was generated, the output was classified as negative~(i.e., the output began with ``No''). For the RoBERTa-large model, we utilized the CLS vector, followed by a linear layer that produced two scores, one for positive and one for negative, to serve as the classifier.

\begin{figure}[h]
    \centering
    \includegraphics[width=0.6\linewidth]{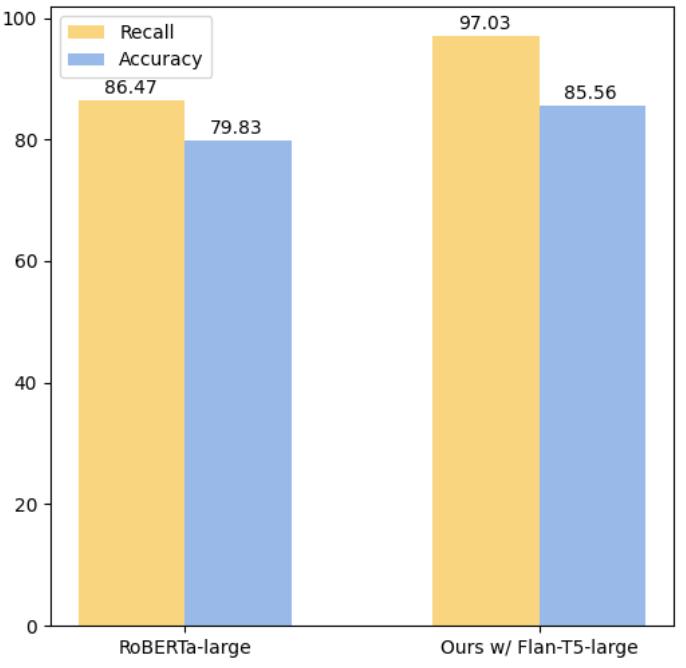}
    \caption{Shared key point detection on ArgKP.}
    \label{fig:shared_key_point_detection}
\end{figure}

We evaluated the recall and accuracy of two models in a binary classification task aimed at identifying whether a pair of arguments shared a key point on ArgKP. As indicated in Figure~\ref{fig:shared_key_point_detection}, our approach outperformed by 10.56 points in recall, the metric of greater significance in our scenario. 
Generating key points may also help in determining whether there are shared key points among arguments.
With pairwise generation, despite the Flan-T5-large being an encoder-decoder model primarily intended for generative tasks, it still delivered satisfactory performance in the classification task.

\subsubsection{SQ2: Key Point Generation}
\begin{figure}[h]
    \centering
    \includegraphics[width=0.85\linewidth]{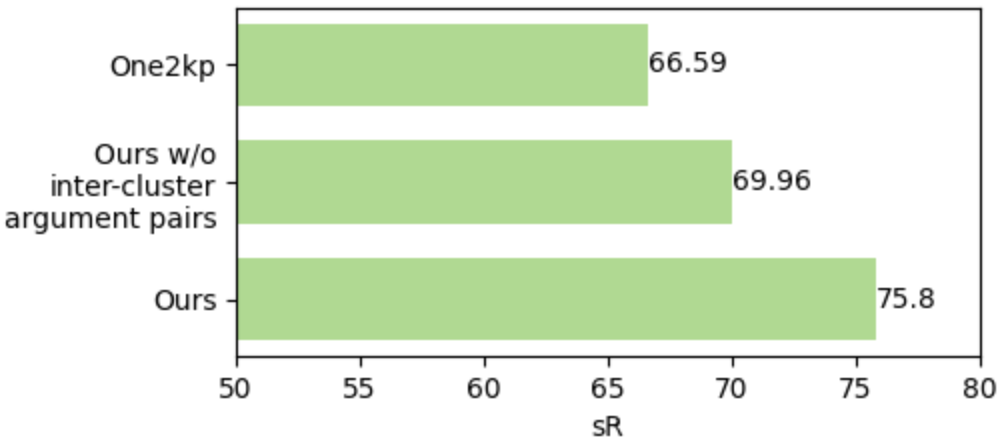}
    \caption{Comparison of different models on key point generation on ArgKP. All the models were based on Flan-T5-large.}
    \label{fig:model_cmp_2}
\end{figure}

In this section, we designed an experiment to verify that the relationship between a pair of intra-cluster arguments with shared key points and inter-cluster arguments without shared key points is beneficial for key point generation. 
We removed the pairs of arguments that did not share key points from our training data as ablation experiment, which is the result of the ``Ours w/o inter-cluster argument pairs'' row in Figure~\ref{fig:model_cmp_2}.
To verify the effectiveness of pairwise generation, we trained a Flan-T5-large model with the concatenation of the topic, stance and a single argument~(i.e., $\texttt{input}=\texttt{topic}\ |\ \texttt{stance}\ |\ \texttt{arg}$) as the input, and an output of concatenated key points implied by the single argument. We denoted this approach as ``One2KP''. 
Here, we also focused on the recall metric~(i.e., sR), which reflected the coverage of the reference key point set during the key point generation stage.

\begin{table*}[th]
    \centering
    \resizebox{\textwidth}{!}{%
    \begin{tabular}{@{\extracolsep{5pt}}lrrrrrrrrrr@{}}
        \specialrule{1pt}{2pt}{2pt}
        & \multicolumn{5}{c}{ArgKP} & \multicolumn{5}{c}{QAM} \\
        \cline{2-6} \cline{7-11} 
        \textbf{Model} & \multicolumn{1}{c}{$\mathtt{Rouge}$-1} & \multicolumn{1}{c}{$\mathtt{Rouge}$-2} & \multicolumn{1}{c}{\textbf{sP}} & \multicolumn{1}{c}{\textbf{sR}} & \multicolumn{1}{c}{\textbf{sF1}} & \multicolumn{1}{c}{$\mathtt{Rouge}$-1} & \multicolumn{1}{c}{$\mathtt{Rouge}$-2} & \multicolumn{1}{c}{\textbf{sP}} & \multicolumn{1}{c}{\textbf{sR}} & \multicolumn{1}{c}{\textbf{sF1}} \\
        \specialrule{1pt}{2pt}{2pt}
        \multicolumn{11}{l}{\textbf{Clustering Methods~(on arguments)}} \\
        k-means$_\text{RoBERTa-large}$ & 50.70 & \underline{21.01} & 59.38 & 59.49 & 59.34  &  29.30 & 7.26 & 48.30 & 44.67 & 46.28 \\
        k-means$_\text{BAAI-bge-large}$ & 49.01 & 18.64 & 60.94 & 60.08 & 60.47 & \underline{30.13} & \underline{7.85} & 47.75 & \underline{44.99} & 46.21  \\
        \specialrule{1pt}{2pt}{2pt}
        \multicolumn{11}{l}{\textbf{Clustering Methods~(on key points)}} \\
        k-means$_\text{RoBERTa-large}$ & \textbf{52.71} & 20.16 & 61.30 & \underline{60.21} & \underline{60.74} & 28.28 & 6.77 & 48.35 & 43.48 & 45.67 \\
        k-means$_\text{BAAI-bge-large}$ & \underline{51.75} & 19.57 & 61.42 & 58.80 & 59.99 & 29.40 & 7.50 & 48.79 & 44.70 & \underline{46.53} \\
        \specialrule{1pt}{2pt}{2pt}
        \multicolumn{11}{l}{\textbf{Graph Partitioning Methods}} \\
        Greedy Modularity Maximization & 40.85 & 14.75 & 56.15 & 54.04 & 55.07 & 18.59 & 4.79 & 39.34 & 33.85 & 36.19 \\
        Louvain Community Detection  & 43.01 & 18.03 & 56.81 & 56.65 & 56.69 & 17.96 & 4.68 & 38.28 & 32.97 & 35.23 \\
        Asynchronous Label Propagation  & 17.32 & 5.54 & \textbf{63.76} & 53.31 & 58.01 & 14.71 & 4.41 & \textbf{54.40} & 40.62 & 46.04 \\
        \specialrule{1pt}{2pt}{2pt}
        \textbf{Ours} & 50.91 & \textbf{21.23} & \underline{62.57} & \textbf{60.26} & \textbf{61.34} & \textbf{30.44} & \textbf{8.02} & \underline{49.38} & \textbf{45.27} & \textbf{47.11}  \\
        \specialrule{1pt}{2pt}{2pt}
    \end{tabular}
    }
    \caption{Comparison with clustering and graph partitioning methods with Flan-T5-large. The best results are in bold and the second best results are underlined.}
    \label{table:effectiness_of_graph_partitioning}
\end{table*}

From Figure~\ref{fig:model_cmp_2}, it can be seen that our method had an advantage over the One2KP method, leading by 9.21 points on the sR metric. After separating inter-cluster argument pairs from the training set, a decrease of 5.84 point on the sR metric was observed. Experiments in Figure~\ref{fig:model_cmp_2} validated that generating key points from pairwise arguments was effective. It was not only necessary to consider the relationships among intra-cluster arguments, but also inter-cluster arguments. 

Additionally, addressing two related tasks simultaneously within the same model did not introduce a drop on the model performance. In our task, as seen in Figure~\ref{fig:shared_key_point_detection} and Figure~\ref{fig:model_cmp_2}, the carefully designed tasks and the use of pairwise generations method brought about a gain in model performance. This allowed our model to not only exhibit superior performance in key point generation but also achieve satisfactory discriminative ability in intra-cluster arguments and inter-cluster arguments classification.

\subsection{Effectiveness of Graph Partitioning}
\label{sec:effectiveness_of_graph_partitioning}
In this section, we aimed to validate the effectiveness of our graph partitioning method. We compared our graph partitioning method with existing clustering methods and graph partitioning methods in Section~\ref{sec:cmp_w_cgpm}. Additionally, we investigated the influence of the number of iterations of our graph partitioning method on the performance in Section~\ref{sec:analysis_of_interations_steps}.

\subsubsection{Comparison with Clustering and Graph Partitioning Methods}
\label{sec:cmp_w_cgpm}
We substituted our graph partitioning algorithm with either clustering methods or graph partitioning methods. 
Our key point generation model, exemplified by the Flan-T5-large model, produced key points for these methods and provided scores to determine if two arguments shared a key point for graph partitioning methods. 
The clustering algorithms we compared were of two types, one clustered arguments to obtain argument subgraphs and then selected key points with highest edge weights, and the other directly clustered key points and then selected the key point at the center of the cluster.
We utilized well-established clustering algorithms, namely k-means~\cite{Lloyd1982LeastSQ} 
, for our analysis. Two models were utilized to derive representations for arguments: 1)~the renowned RoBERTa-large~\cite{Liu2019RoBERTaAR} and 2)~BAAI-bge-large~\cite{Xiao2023CPackPR}, which is the state-of-the-art model in sentence representation. Furthermore, we employed three well-known graph partitioning methods, including
\begin{itemize}
    \item Greedy Modularity Maximization~\cite{Clauset2004FindingCS}, which is a graph partitioning algorithm that utilizes a greedy strategy to maximize modularity and identify community structures,
    \item Louvain Community Detection~\cite{Blondel2008FastUO}, which is a heuristic method based on modularity optimization by iteratively merging similar vertices to uncover community structures, and
    \item Asynchronous Label Propagation~\cite{PhysRevE.76.036106}, which relies on label propagation iterations to divide the graph into communities by updating vertex labels.
\end{itemize}

From Table~\ref{table:effectiness_of_graph_partitioning}, our model exhibited superior performance compared to the three sets of baseline models on $\mathtt{Rouge}$-2, sR and sF1 metrics. Specifically, on the ArgKP dataset, our model exhibited a lead of 0.60--6.27 points in sF1. On the QAM dataset, our model demonstrated a lead of 0.58--11.88 points in sF1.

Our approach amalgamated the strengths of both by identifying shared key points through our generative model, taking into account semantic information, while simultaneously utilizing graph structure information via the graph partitioning method, resulting in improved outcomes.

\subsubsection{Analysis of Iterations Steps~($l$)}
\label{sec:analysis_of_interations_steps}

\begin{table}[h]
    \centering
    \small
    \resizebox{0.9\linewidth}{!}{%
    \begin{tabular}{@{\extracolsep{5pt}}lrrrrr@{}}
        \specialrule{1pt}{2pt}{2pt}
        \textbf{\#steps} & $\mathtt{Rouge}$-1 & $\mathtt{Rouge}$-2 & \textbf{sP} & \textbf{sR} & \textbf{sF1} \\
        \specialrule{1pt}{2pt}{2pt}
        0 & 49.01 & 18.64 & 60.94 & 60.08 & 60.47 \\
        50 & 51.46 & 20.55 & 61.38 & 60.63 & 60.96 \\
        100 & 47.53 & 19.77 & 60.92 & 59.25 & 60.03 \\
        150 & 52.13 & 22.39 & 61.22 & 60.52 & 60.85 \\
        200 & 50.91 & 21.23 & 62.57 & 60.26 & 61.34 \\
        250 & 49.23 & 19.45 & 62.32 & 59.93 & 61.09 \\
        \specialrule{1pt}{2pt}{2pt}
    \end{tabular}
    }
    \caption{Analysis of iterations steps~($l$).}
    \label{table:analysis_of_iterations_steps}
\end{table}

We conducted an analysis on the impact of the number of iterations steps in the graph partition algorithm. When the number of steps equaled 0, it represented a ablation experiment, indicating the exclusion of any execution of the graph partition algorithm. As shown in Table~\ref{table:analysis_of_iterations_steps}, it could be seen that, excluding the subtle fluctuations in the data, the overall trend of sF1 was first an increase and then a stabilization. Additionally, it could be seen that as the algorithm iterations progress, sR did not change much overall, and our graph partitioning algorithm improved the overall performance by improving sP.

\section{Related Work}
\label{sec:related_work}
\subsection{Key Point Analysis}
Unlike multi-document summarization~\cite{Mani1997MultiDocument,xiao-etal-2022-primera}, KPA~\cite{bar-haim-etal-2020-arguments} requires the production of a concise collection of key points rather than a cohesive paragraph. Furthermore, KPA entails a quantitative analysis of these key points, involving quantifying their prevalence within the arguments. 

Existing methods for key points generation in KPA rely on extractive summarization and abstractive summarization techniques.
Extractive summarization methods~\cite{bar-haim-etal-2020-quantitative,bar-haim-etal-2021-every,alshomary-etal-2021-key} involve selecting short and representative arguments as key points. However, extractive summarization approaches are limited by the quality of the arguments and do not ensure the presence of an optimal key point within the arguments. 
On the other hand, abstractive summarization methods utilize generative models to produce key points for the arguments. \citealp{kapadnis-etal-2021-team} generates key points for each argument individually and then select a collection of key points based on the Rouge~\cite{lin-2004-rouge} scores.
\citealp{li-etal-2023-hear} first employs clustering methods to group semantically similar arguments into the same cluster, and then concatenates the arguments within the cluster to generate key points.

In contrast, our approach simultaneously takes into account the relationships between arguments with shared key points and those without shared key points. Additionally, by replacing semantic similarity scores with scores for the presence of shared key points between arguments, we address the shortcomings of existing models.

\subsection{Graph Partitioning}
Unlike clustering methods which depends on the similarity or distance of clustering objects, graph partitioning aims to optimize a certain metric of the resulting subgraphs~\cite{Bulu2013RecentAI}, which is the probability that the arguments within a common subgraph share a common key point in the KPA scenario.
The algorithms~\cite{6771089,10.5555/1496770.1496872,10.14778/2824032.2824046} based on optimizing the minimum vertex/edge cuts has been extensively utilized. Additionally, algorithms~\cite{Clauset2004FindingCS,Blondel2008FastUO,Dugu2015DirectedL,Traag2018FromLT} have approached graph partitioning by optimizing the modularity of subgraphs. Another avenue~\cite{PhysRevE.76.036106,5730298} involves achieving graph partitioning through vertex labels propagation. 

In the KPA task, our objective is to maximize the probability that arguments within a subgraph share the same key point while allowing arguments that contain more than one key point to exist in multiple subgraphs simultaneously. Given this requirement, the aforementioned graph partitioning algorithms are not suitable for the KPA task.

\section{Conclusion}
\label{sec:conclusion}

This paper presents a method that merges pairwise generation with graph partitioning to efficiently produce key points within arguments. Our integrated approach models both the relationships between arguments sharing key points and those that do not, aiding in key point generation. We employ an iterative algorithm to partition the argument graph and extract representative key points from subgraphs, yielding a concise set of key points.  In the future, we will continue to explore better ways to utilize generative models to model the relationships among arguments. For instance, we will investigate improved methods of outputting scores indicating whether arguments share key points and develop graph partition algorithms that better align with generative models.
Our approach may provide inspiration for tasks including clustering tasks based on generative models~\cite{Zhang2023ClusterLLMLL,Viswanathan2023LargeLM}, set generation tasks~\cite{madaan-etal-2022-conditional}, and so forth.

\section*{Limitations}
The input consists of pairwise arguments, which increases the amount of training data and the computational cost of training. Currently, training time of Flan-T5-large~(780M) on ArgKP is approximately 12 hours, and on QAM it is about 6 hours. Our model does not fully utilize the graph structure during the graph partitioning process, and we plan to incorporate more graph structural information into the model in the future. The datasets currently used for experiments are related to survey review topics, and there is a lack of datasets from other fields~(such as product reviews in the e-commerce scenario) in the research community. Future research could explore the effectiveness on datasets from different domains.


\bibliography{refs}


\end{document}